\documentclass{article}

\PassOptionsToPackage{numbers}{natbib}
\usepackage[preprint]{neurips_2021}

\usepackage[utf8]{inputenc} 
\usepackage[T1]{fontenc}    
\usepackage{booktabs}       
\usepackage{amsfonts}       
\usepackage{nicefrac}       
\usepackage{microtype}      
\usepackage{amsmath}
\usepackage{graphicx}
\usepackage{tipa}
\usepackage{multirow}
\usepackage{longtable}

\usepackage{subcaption}
\usepackage{color}
\usepackage[table]{xcolor}
\usepackage{threeparttable}
\usepackage{diagbox}
\usepackage{enumitem}
\usepackage{tablefootnote}
\usepackage{hyperref}
\usepackage{url}            
\usepackage{xurl}
\usepackage{breakurl}

\usepackage[edges]{forest}
\usepackage{tikz}
\usepackage{tikz-qtree}
\usetikzlibrary{fadings}
\usetikzlibrary{mindmap,trees}
\usetikzlibrary{arrows,automata,shapes,positioning,shadows,trees}
\usetikzlibrary{shapes,snakes,shadows}
\usetikzlibrary{shapes.arrows}
\usetikzlibrary{calc,shapes, positioning}
\usetikzlibrary{decorations.text}
\usetikzlibrary{matrix,chains,positioning,decorations.pathreplacing,arrows}
\usetikzlibrary{bayesnet}
\usetikzlibrary{shadows.blur}
\usetikzlibrary{shapes.geometric}

\tikzstyle{edge}=[-latex',draw=black!90,shorten <=1pt,shorten >=1pt]
\tikzstyle{redge}=[latex'-,draw=black!90,shorten <=1pt,shorten >=1pt]
\tikzstyle{dedge}=[latex'-latex',draw=black!90,shorten <=1pt,shorten >=1pt]
\tikzstyle{block}=[draw, text width=5em,align=center,shape=rectangle, rounded corners, , align=center]
\tikzstyle{nobox}=[align=center]

\definecolor{emb}{RGB}{209,228,252}
\definecolor{hidden-blue}{RGB}{194,232,247}
\definecolor{hidden-orange}{RGB}{243,202,120}
\definecolor{hidden-yellow}{RGB}{242,244,193}
\definecolor{output-purple}{RGB}{219,203,231}
\definecolor{output-green}{RGB}{204,231,207}
\definecolor{hiddendraw}{RGB}{205, 44, 36}

\tikzstyle{mybox}=[
    rectangle,
    draw=hiddendraw,
    rounded corners,
    text opacity=1,
    minimum height=1.5em,
    minimum width=5em,
    inner sep=2pt,
    align=center,
    fill opacity=.2,
    ]

\tikzstyle{emb-purple}=[
    rectangle,
    draw=output-purple!50!purple,
    fill=output-purple,
    text opacity=1,
    minimum height=1.5em,
    minimum width=1.5em,
    inner sep=0pt,
    align=center,
    fill opacity=.9,
    ]

\tikzstyle{emb-blue}=[
    rectangle,
    draw=emb!50!blue,
    fill=emb,
    text opacity=1,
    minimum height=1.5em,
    minimum width=1.5em,
    inner sep=0pt,
    align=center,
    fill opacity=.9,
]

\definecolor{colone}{RGB}{178, 34, 34}
\definecolor{coltwo}{RGB}{106, 90, 205}
\definecolor{colthree}{RGB}{255, 250, 205}
\definecolor{colfour}{RGB}{0, 139, 69}
\definecolor{colfive}{RGB}{245,238,197}
\definecolor{colsix}{RGB}{243,235,179}
\definecolor{colseven}{RGB}{241,231,163}
\title{A Survey on Image Deblurring}

\author{
ChuMiao Li\\
\texttt{xihuamastermiao@163.com} \\
Xihua University\\
}

\begin{document}
\maketitle

\begin{abstract}

With the improvement of social life quality and the real needs of daily work, images are more and more all around us. Image blurring due to camera shake, human movement, etc. has become the key to affecting image quality. How to remove image blur and restore clear image has gradually become an important research direction in the field of computer vision. After more than half a century of unremitting efforts, the majority of scientific and technological workers have made fruitful progress in image deblurring. This article reviews the work of image deblurring and specifically introduces more classic image deblurring methods, which is helpful to understand current research and look forward to future trends. This article reviews the traditional image deblurring methods and depth-represented image deblurring methods, and comprehensively classifies and introduces the corresponding technical methods. This review can provide some guidance for researchers in the field of image deblurring, and at the same time facilitate their subsequent study and research.

\end{abstract}

\section{introduction}
\label{inrtoduction}
Image deblurring aims to restore blurred pictures into clearer pictures, and has a wide range of application scenarios in human society and industrial production. Because image blurring is widespread in life, the problem of image deblurring has received attention and research since the last century. Since the 1960s, in order to solve the problem of image deblurring, it was first proposed to transform the image problem into the frequency domain to solve it, and classical algorithms such as inverse filtering~\cite{2006Image} and Wiener filtering~\cite{2007Image} were proposed. However, the image deblurring algorithm based on the frequency domain needs to know the degradation type of the blur accurately and is sensitive to noise, and then an estimation algorithm based on the spatial domain is proposed. Common estimation algorithms based on the spatial domain include: differential restoration algorithm~\cite{1989A}, least square algorithm~\cite{2004The}, maximum entropy algorithm~\cite{0Passive} and so on. In recent years, the image deblurring algorithm has made greater breakthroughs. Over time, many methods and categories have been derived from the problem of deblurring. According to the method of deblurring algorithm, classification can be roughly divided into optimization-based deblurring method and deep learning deblurring method; classification according to input can be divided into: deblurring algorithm based on single frame input and deblurring based on multi-frame input Fuzzy algorithm; classification according to the degradation type can be divided into motion blur, Gaussian blur, defocus blur, turbulence blur, etc.; classification according to whether the blur kernel is known can be divided into blind deblurring and non-blind deblurring; according to whether the degradation is Consistency can be divided into space-changing blur and space-invariant blur. 

Image deblurring is a classic problem in computer vision research. In recent years, image deblurring has become a basic topic in the field of computer vision, and it is also one of the key projects that can be applied. As shown in Figure 1, the so-called image deblurring refers to the restoration of the extreme visual experience based on the given blurred picture with poor look and feel.The process of good clear pictures. In recent years, with the rise of deep learning technology, the effect of image deblurring has made a qualitative leap. Its applications are mainly focused on space exploration, astronomical observation, material research, remote sensing and telemetry, military science~\cite{2014An}, biological science, and medical imaging.~\cite{2011Alternating}, traffic monitoring~\cite{2012A}, criminal investigation and other fields.

\begin{figure}[htbp]
\begin{minipage}[t]{0.5\linewidth}
\centering
\includegraphics[height=3cm,width=4cm]{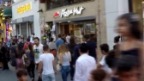}
\subcaption{Blurred image}
\end{minipage}
\begin{minipage}[t]{0.5\linewidth}
\centering
\includegraphics[height=3cm,width=4cm]{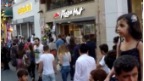}
\subcaption{Clear image}
\end{minipage}
\caption{Schematic diagram of image deblurring}
\end{figure}

The image blur process can be simulated as $b = Ks + n$, where $b$, $s$, and $n$ represent the vectorized blurred image, the clear latent image, and the additional noise, respectively. Therefore, based on the traditional deblurring method, recovering a clear picture under the condition of a known blurred picture is a highly unsuitable problem. However, the blurring can be removed by applying different constraints to the blur kernel and the latent image, or using hand-made regularization items at the same time~\cite{2021Multi}. Unfortunately, these methods often lead to solving a non-convex optimization problem, so they cannot be well generalized to complex real examples~\cite{2021BANet}. In recent years, in~\cite{2020EDGAN,2016Densely,2019DeblurGAN,2020Deep,2020Strip,2020Learning,2020Cascaded,2019Region,2020Deblurring,2020Efficient,2018Scale,2019Deep,2019Dynamic,2020Multi,2017Non,2015Learning}, the method based on learning representation has achieved great success in the field of image deblurring, and its objective indicators and subjective feelings have improved significantly. The deblurring network with a large number of learning samples can accurately estimate the fuzzy kernel and recover a clear picture, and this end-to-end model is easy to understand and easy to apply.

In general, this paper mainly introduces two main methods of image deblurring and general data sets for image deblurring. In Section 2, we mainly introduce traditional image deblurring methods, and in Section 3, we mainly introduce image deblurring with depth representation. Methods and their ideas. In Section 4, we mainly introduce several commonly used image deblurring datasets, and in Section 5, we mainly introduce image deblurring evaluation indicators. The main significance of this review is:
\begin{itemize}[leftmargin=*]
    \item Introduced and summarized several more classic traditional methods and depth representation methods for image deblurring tasks.
    \item This review provides some guidance for researchers interested in this direction.
\end{itemize}

\tikzstyle{leaf}=[mybox,minimum height=1.2em,
fill=hidden-orange!50, text width=5em,  text=black,align=left,font=\footnotesize,
inner xsep=4pt,
inner ysep=1pt,
]

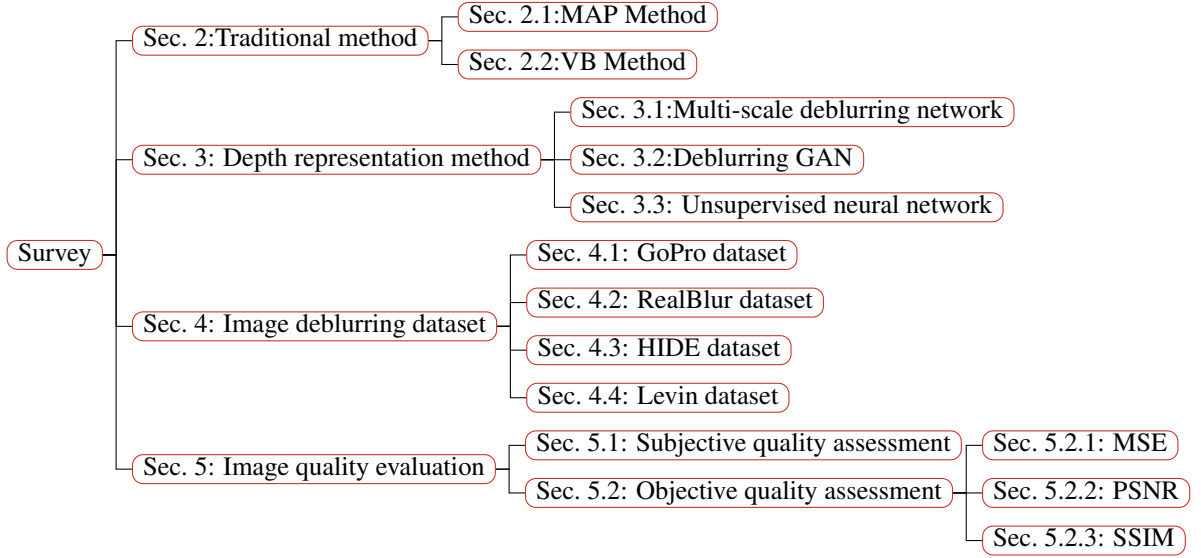
\begin{figure*}[thp]
 \centering
\begin{forest}
  forked edges,
  for tree={
  grow=east,
  reversed=true,  
  anchor=base west,
  parent anchor=east,
  child anchor=west,
  base=left,
  font=\normalsize,
  rectangle,
  draw=hiddendraw,
  rounded corners,
  align=left,
  minimum width=2.5em,
  inner xsep=4pt,
  inner ysep=0pt,
  },
    [Survey
        [Sec.~\ref{Traditional image deblurring method}:Traditional method
            [Sec.~\ref{Maximum posterior method}:MAP Method      
            ]
            [Sec.~\ref{Variational Bayes Method}:VB Method
            ]
        ]
        [Sec.~\ref{Depth representation method}: Depth representation method
            [Sec.~\ref{Multi-scale deblurring network}:Multi-scale deblurring network 
            ]
            [Sec.~\ref{Generative adversarial deblurring network}:Deblurring GAN 
            ]
            [Sec.~\ref{Unsupervised neural network}: Unsupervised neural network
            ]
        ]
        [Sec.~\ref{Image deblurring dataset}: Image deblurring dataset
            [Sec.~\ref{GoPro dataset}: GoPro dataset
            ]
            [Sec.~\ref{RealBlur dataset}: RealBlur dataset
            ]
            [Sec.~\ref{HIDE dataset}: HIDE dataset
            ]
            [Sec.~\ref{Levin dataset}: Levin dataset
            ]
        ]
		[Sec.~\ref{Image quality evaluation}: Image quality evaluation
            [Sec.~\ref{Subjective quality assessment}: Subjective quality assessment
            ]
            [Sec.~\ref{Objective quality assessment}: Objective quality assessment
				[Sec.~\ref{MSE}: MSE
            	]
				[Sec.~\ref{PSNR}: PSNR
            	]
				[Sec.~\ref{SSIM}: SSIM
            	]
			]
        ]
    ]
\end{forest}
\caption{Organization of this paper.}
\label{org_survey_paper}
\end{figure*}

\section{Traditional image deblurring method}
\label{Traditional image deblurring method}
Early classic image deblurring algorithms are mainly Wiener filtering algorithm~\cite{2007Image}, RL filtering algorithm, total variation algorithm, etc. Because these methods fail to make full use of the prior information of natural images, the restored image is not accurate and mainly used Deblurring for non-blind images. In view of the shortcomings of the classic deblurring methods, the later traditional methods make full use of some prior information of the natural image to restore the image based on the classic method, such as the sparsity of the natural image, and overcome the defects of the original method. Generally, the prior information of natural images is mainly used to obtain the prior information related to natural images through a large number of statistical image characteristics, and use it as a regularization item to improve the deblurring effect of the image.

The model-based method is to model the fuzzy process first, and then solve the inverse process with the help of mathematical tools such as optimization methods. As mentioned earlier, the classic forms of traditional deblurring are:
\begin{equation}\label{eq1}
y=k*x+n
\end{equation}
Among them, $n$ represents noise. We know the blurred image $y$, and then estimate the original image $x$ and the blur kernel $k$ (blind deblurring), or know $y$ and $k$ to estimate $x$ (unblind deblurring). The solution of the inverse process is often an ill-conditioned problem, and the solution to this problem is often not unique.
\begin{equation}\label{eq2}
x,k=\underset{x,k}{argmin}\left [ \lVert x\otimes k-y \rVert_{2}^{2}+\alpha \left ( x \right )+\beta \left ( k \right ) \right ]
\end{equation}
Later, a method based on regularization was proposed. The method based on regularization technology is to construct an optimized model as shown in equation \ref{eq2}, so that the restored image can make the above equation take the minimum value. Among them, $x$ represents the final restored image, and $k$ represents the estimated point spread function. Equation \ref{eq2} The first term on the right side of the equal sign is called the fidelity term, which is used to ensure that the restored image can be consistent with the original blurred image. The second term $\alpha \left ( x \right )$ is called the regularization term of the restored image. The function of this term is to make the restored image meet a certain condition. This term is a function of the restored image $x$. The third term $\beta \left ( k \right )$ is a regularization term about the point spread function, and its function is to make the estimated point spread function meet a certain condition. When the regularization term on the point spread function exists, this type of algorithm estimates the point spread function and the clear image at the same time. This type of algorithm is also called blind deblurring algorithm or blind deconvolution algorithm. When it is assumed that the point spread function is known, it is only necessary to estimate a clear image that minimizes the value of the image deblurring mathematical model. The change of the mathematical model of image deblurring is expressed in the following form:
\begin{equation}\label{eq3}
x=\underset{x}{argmin}\left [ \lVert x\otimes k-y \rVert_{2}^{2}+\alpha \left ( x \right ) \right ]
\end{equation}
Such algorithms are also called non-blind deblurring algorithms or non-blind deconvolution algorithms. The construction of the regularization term is the key part of image deblurring based on regularization. The regular term of the point spread function and the regular term of the restored image will have an important influence on the restoration result.
\subsection{Maximum posterior method}
\label{Maximum posterior method}
In order to reduce the solution space of the problem and better approximate the real solution, we need to add prior conditions. Different a priori conditions will lead us into different solution spaces, and thus get different restored images. In the model-based method, the objective function of image deblurring is established using the method of maximum posterior probability under the Bayesian framework. However, model-based deblurring methods also have their limitations. The first is that most methods are based on the classic convolution model, aiming at globally uniform blur. For non-global uniform blur, such as image rotation blur caused by camera rotation around the Z axis, it is difficult to use a single blur kernel to describe. It is even more helpless for the local motion blur caused by the motion of objects in the scene. In addition, model-based methods generally use an iterative approach, using image pyramids to estimate blur kernels and remove blurs from small to large, which often requires a lot of time and overhead. Professor Levin believes that it is definitely not good to use the MAP objective function to solve k and x at the same time. There are the following reasons: First, the wrong unit convolution kernel is more likely than the correct sparse convolution kernel (proved in Levin’s paper ); Secondly, the number of variables of the above-mentioned MAP objective function has a high degree of asymmetry, which means that the number of known quantities is always smaller than the number of unknown quantities. The above two reasons will lead to the use of this objective function of simultaneously estimating k and x to do MAP always can not get good results. How to solve this asymmetric problem is the key to breaking the game. Levin believes that a better way is to estimate k alone, so that the number of known quantities is much larger than the number of unknowns. To put it simply, for each possible k, we have to find the posterior probability of all possible x, and add these posterior probability values to get $p\left ( k|y\right )$.

In ~\cite{2008High},the author found in experiments that inaccurately modeled image noise and errors in the estimated point spread equation are the main causes of ringing artifacts, not the Gibbs phenomenon previously thought.The author thinks Gradient-based optimization is the most feasible way to optimize such a log-prior, and gradient-descent is known to be neither effificient nor stable for a complex energy function containing thousands or millions of unknowns.The author proposes a new representation by concatenating two piece-wise continuous functions to fifit the logarithmic gradient distribution.Secondly,the author use the blurred image to constrain the gradients of the latent image in a fashion that is very effective in suppressing ringing artifacts. a new smoothness constraint that we impose on the latent image in areas where the observed image has low contrast. This constraint is very effective in suppressing ringing artifacts not only in smooth areas but also in nearby textured ones. The effects of this constraint propagate to the kernel refifinement stage, as well.The author combines the above-mentioned global image prior and local image prior to form an image prior.

In ~\cite{2007Single},the author proposes a unified method for estimating the motion blur filter from the perspective of transparency.At the same time,the author propose an optimization method to estimate the blur fifilter by solving a maximum a posteriori (MAP) problem only using transparency information. Without directly taking all pixel colors and complex image structures into computation, this method is effificient and robust. The transparency defifinition in a camera motion blurred image regarding the region blending is different from that on a motion blurred object. Assuming that the moving object occludes static background behind, in the object motion blur, the transparency map can be uniquely determined.The new transparency map can be defifined differently, so the author call it the generalized transparency. The author uses this prior knowledge of clarity to solve the maximum posterior probability of the fuzzy kernel.

In ~\cite{2009Image},the author propose an additional term in the sparse prior that uses a color model built from local color statistics of the sharp latent image. This overcomes over-smoothing as it allows for sharp edges as long as they are consistent with local color statistics.In contrast with a gradient prior, which prefers the lowest intensity edges that are consistent with the observed blurred image, a two-color model can result in higher intensity edges if such edges are more consistent with local color statistics.

In ~\cite{2013Blind},The novelty is to use image priors $P\left ( u \right )$ that are more heavy-tailed than Laplace distribution and apply a method of augmented Lagrangian to tackle this non-convex optimization problem.

In ~\cite{2011Blind},the regularization function the author propose is the ratio of the l1-norm to the l2-norm on the high frequencies of an image,that can distinguish between blurred images and clear images very well.At the same time, it does not require the complexity required by other methods to overcome the shortcomings of the existing priors in the MAP setting to estimate the blur kernel and clear pictures.

\begin{figure}[htbp]
\centering
\includegraphics[height=12cm,width=10cm]{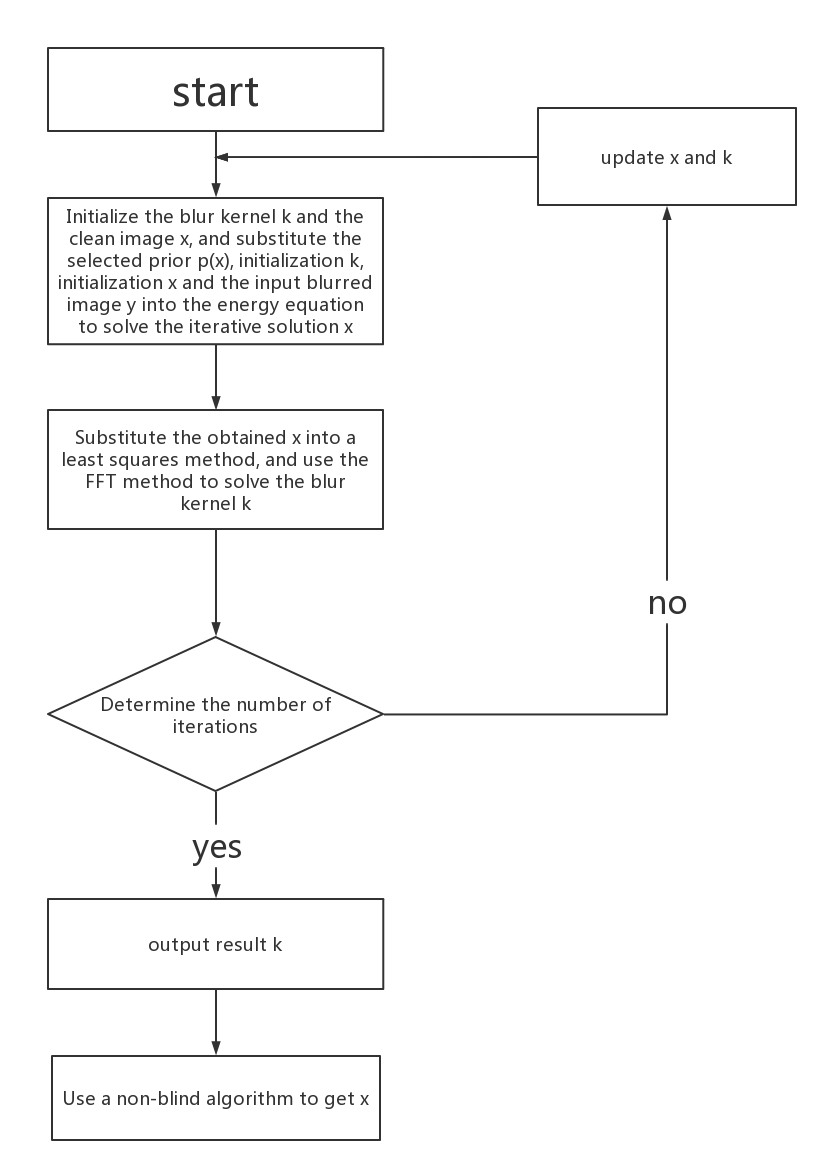}
\caption{MAP method flow chart}
\end{figure}

\subsection{Variational Bayes Method}
\label{Variational Bayes Method}
It is not easy to find the above marginal probability, and the amount of calculation is very complicated. So many scholars have adopted an approximate method to solve the problem. Fergus~\cite{2006Removing} et al. proposed the first non-parametric motion blur kernel estimation method in 2006, which is also the earliest blind image deblurring method. By learning the mixed Gaussian image prior and the mixed exponential fuzzy kernel prior, the author proposes an image non-parametric deconvolution method based on the VB criterion. In the past, the estimation of fuzzy kernel based on MAP method mostly failed. Although the actual deblurring effect of this method is not satisfactory, the enlightening significance brought by it is far-reaching. On the one hand, using the posterior point estimation criterion of MAP (which is largely equivalent to the regularization method), the mixed Gaussian image prior is completely unsuitable for the blind deblurring problem, which is not conducive to the effective identification and estimation of saliency edges; another On the one hand, compared with the point estimation of MAP, the posterior mean estimation of VB has natural robustness, which can reduce the risk of "significant edge-non-parametric fuzzy kernel" alternate iteration estimation falling into a local minimum~\cite{wipf2014revisiting}. In addition to the enlightening elaboration of this article, the follow-up literature~\cite{He2016Deep,2016Deep,2016Photo,2016Image}has proved the superiority of the image deblurring method based on VB compared with the MAP method from the perspective of theory and practice. In the paper, Fergus showed its complete blind deconvolution process: First, preprocess the image. In order to reduce the amount of calculation and get a good result, the user is required to select an image block; secondly, the use of variational Bayes , Estimate the convolution kernel k. In order to avoid falling into the local optimum, the author adopts the coarse-to-fine strategy; finally, the standard non-blind deconvolution method is used to reconstruct the clear image x.

In ~\cite{2006Removing},the author found that recent research in natural image statistics, which shows that photographs of natural scenes typically obey very specifific distributions of image gradients.The author uses a zero-order Gaussian mixture model to represent the distribution of the gradient size. The reason for choosing this representation is that it can provide a good approximation of the empirical distribution and allow the algorithm to have a processable estimation procedure.The author believes that the objective function of MAP tries to minimize all gradients (even larger gradients), and we expect natural images to have some larger gradients. In addition, the author found that the MAP objective function is easily affected by local minima.

In ~\cite{2009Understanding},the author further explored the reasons for the failure of the MAP method, and concluded in the $MAP_{x,k}$ problem we can never collect enough measurements because the number of unknowns grows with the image size.The author believes that the number of variables of the MAP objective function has a high degree of asymmetry. It can be seen that the number of known quantities is always less than the number of unknowns.At the same time, the author proved for an increasing image size, a $MAP_{k}$ estimation of k alone (marginalizing over x) can recover the true kernel with an increasing accuracy.Moreover, the method of obtaining the variational Bayesian approximation through quantitative analysis is indeed significantly better than the existing methods.

In ~\cite{wipf2014revisiting},the author systematically analyzed and demonstrated the basic principles of the VB method and the MAP method,demonstrated that rigorous evaluation of VB and its associated priors cannot be separated from implementation heuristics, and we have meticulously examined the interplay of the relevant underlying algorithmic details employed by practical VB systems.At the same time, it further analyzes the priori optimal selection problem of the VB blind deblurring method theoretically and points out that to some extent, Jeffreys prior can achieve the best blind deblurring effect.

\tikzstyle{leaf}=[mybox,minimum height=1.2em,
fill=hidden-orange!50, text width=5em,  text=black,align=left,font=\footnotesize,
inner xsep=4pt,
inner ysep=1pt,
]

\begin{figure*}[thp]
 \centering
\begin{forest}
  forked edges,
  for tree={
  grow=east,
  reversed=true,  
  anchor=base west,
  parent anchor=east,
  child anchor=west,
  base=left,
  font=\normalsize,
  rectangle,
  draw=hiddendraw,
  rounded corners,
  align=left,
  minimum width=2.5em,
  inner xsep=4pt,
  inner ysep=0pt,
  },
    [Traditional deblurring method
        [Sec.~\ref{Maximum posterior method}: Maximum posterior method
        	[Qi et al.~\cite{2008High}
			]
			[Jia et al.~\cite{2007Single}
			]
			[Joshi et al.~\cite{2009Image}
			]
			[Kotera et al.~\cite{2013Blind}
			]
			[Krishnan et al.~\cite{2011Blind}
			]
        ]
        [Sec.~\ref{Variational Bayes Method}:Variational Bayes Method 
			[Fergus et al.~\cite{2006Removing}
			]
			[Levin et al.~\cite{2009Understanding}
			]
            [Wipf et al.~\cite{wipf2014revisiting}
            ]
        ]
    ]
\end{forest}
\caption{Classification based on traditional image deblurring methods}
\label{traditional_deblurring_classification}
\end{figure*}
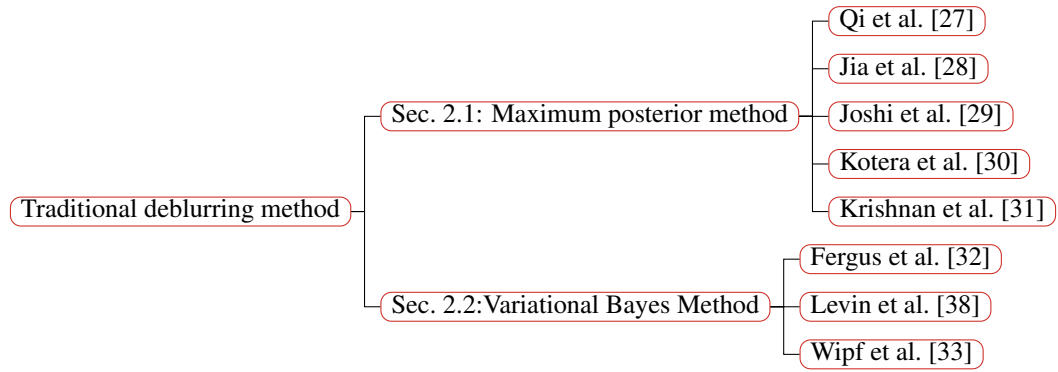

\section{Depth representation method}
\label{Depth representation method}
With the rapid development of deep learning in recent years, using a data-driven approach to solve the problem of blind deblurring of images seems to be another normalized method after VB and MAP methods. Simply put, the focus of current data-driven blind deblurring methods often lies in how to build a supervised deep neural network with good generalization performance. It is obvious that the performance of this method is mainly different from the rationality of the convolutional neural network architecture design and the rationalization of training. Specifically, many highly representative models emerged in the ImageNet ILSVRC Image Recognition Challenge, which largely reflects the relevant progress in the early stage of the development of Deep Representation, including: AlexNet, ZFNet, GogleNet, VGGNet, ResNet, ResNeXt, etc. The famous international academic conferences such as ICCV, CVPR, ECCV, etc. successively held different computer vision tasks such as facial recognition, object detection, semantic segmentation, etc., also provide a new idea for the network construction of image deblurring. As far as the current development trend is concerned, we mainly introduce multi-scale de-fuzzing networks and the relatively hot generation of confrontation de-fuzzing networks and unsupervised de-fuzzing networks in recent years.

Jian Sun et al.~\cite{2015Learning}first applied the deep learning method to blind image deblurring, and proposed a deep learning method to predict the probability distribution of motion blur at the patch level using convolutional neural networks. The author uses a well-designed image rotation to further expand the set of candidate motion blur kernels predicted by CNN. Then use the Markov random field model to infer a dense non-uniform motion blur field, thereby enhancing the smoothness of the motion. Finally, the patch-level image prior non-uniform deblurring model is used to remove motion blur.

Ayan Chakrabarti et al.~\cite{2016A}applied the complex Fourier coefficients of the output deconvolution filter of the training network to the input patch, and used multi-resolution frequency decomposition to encode the input patch, and based on the locality in frequency (similar to the receiving space Convolutional layer with locality restriction) to limit the connectivity of the initial network layer. This leads to a significant reduction in the number of weights that need to be learned during the training process, which proves to be crucial because it allows us to successfully train a network that runs on large patches and can therefore explain large fuzzy kernels.
\subsection{Multi-scale deblurring network}
\label{Multi-scale deblurring network}
Multi-scale network design has been very useful in other areas of image tasks. For example, the article ~\cite{2019A2RMNet}and~\cite{2021A}, the idea of multi-scale is applied to the field of target detection, such as the ~\cite{Luo2020Multi}and~\cite{2021Single}, which puts the idea of multi-scale into practice In the work of removing rain from images, such as~\cite{2019A}, the multi-scale method has been successfully applied to the field of image segmentation. The success of multi-scale thinking in other image tasks makes it natural to migrate to the field of image deblurring.

\begin{figure}[htbp]
\centering
\includegraphics[height=10cm,width=10cm]{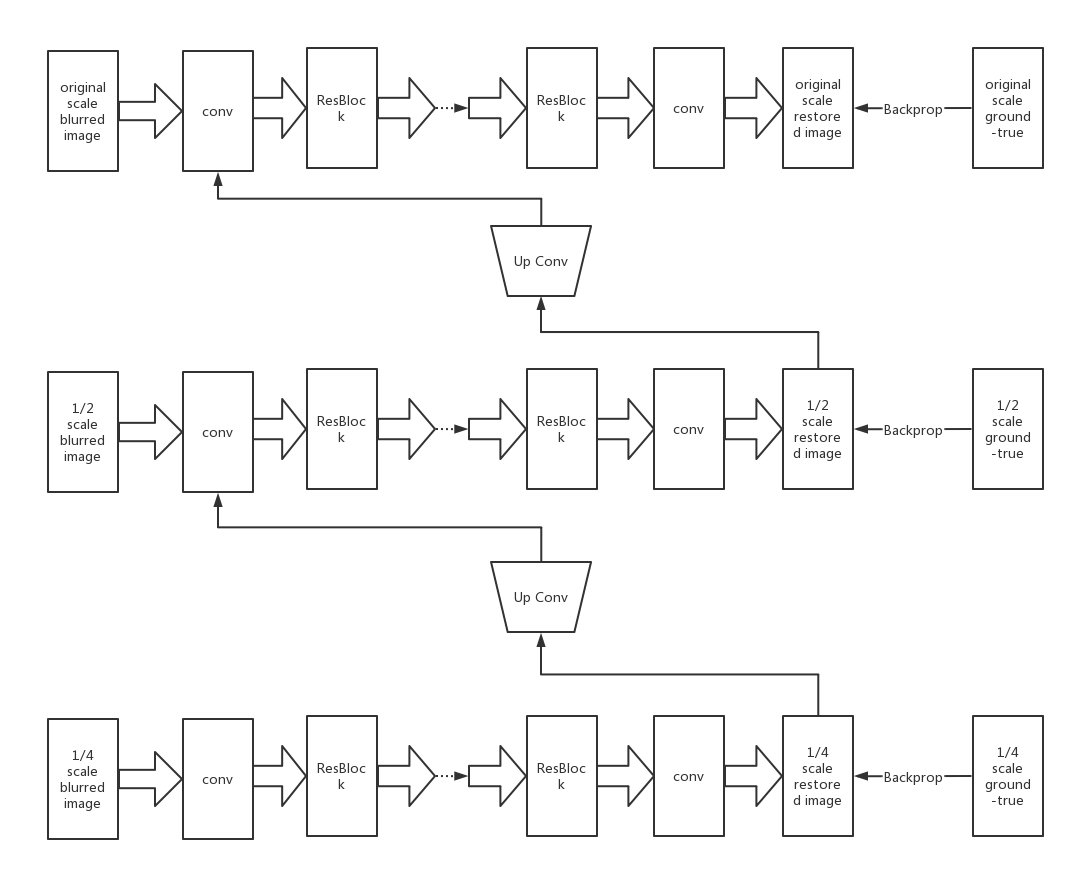}
\caption{Multiscale Model Framework Diagram}
\end{figure}

Seungjun et al.~\cite{2016Deep} based on deep learning methods and traditional optimization methods have proposed to solve the problem of non-uniform blur of the image, but for the blur problem of dynamic scenes, that is, the problem of blurring only partial areas in the image is difficult. solve. Through the analysis and summary of the previous algorithms, the following problems are summarized: First, for the data set, it is difficult to obtain the measured clear and blurred images for training; second, for the deblurring problem in dynamic scenes, it is difficult to obtain the blur of the partial images. Nuclear; Finally, a larger receptive field helps to solve the problem of image deblurring. In order to simulate the traditional optimization method from coarse to fine, the author designed a multi-scale deblurring model. As the name suggests, in the first stage, the author downsampled the original scale of the blurred image to 1/4 of the original size and input it into the network to obtain the largest restoration result; through the up-convolution operation, the result obtained at the 1/4 scale stage was compared with the original The fuzzy images of the 1/2 scale of the image are combined as the input of the second stage, and the deblurring result of the second stage is obtained through the network; the output of the second stage is convolved and combined with the original size of the blurred image as the input, Get the finest clear image through the network model. This multi-scale strategy is similar to the coarse-to-fine strategy in traditional fuzzy kernel estimation. It decomposes complex problems and restores them step by step. First, restore large-scale information at low resolution, and then high resolution. Recover detailed information. This design simplifies the problem while increasing the field of perception of the image.

Later, Xin Tao et al.~\cite{2018Scale} proposed that because each scale solves the same problem, the same model parameters can be used. This cross-scale sharing of network weights can significantly reduce the difficulty of training, while introducing obvious stability benefits. Such a design significantly reduces the number of trainable parameters, and the recycling of shared weights works similarly to using data multiple times to learn parameters.

Hongyun Gao et al.~\cite{2019Dynamic} found that not all module parameter sharing is conducive to the improvement of defuzzification performance, and thus proposed the general principle of selective parameter sharing, which can be conducive to the design of future defuzzification systems. Since in resnet~\cite{2016Deep} and reblocks~\cite{He2016Deep}, the short-term jump connection of adding the input to the output after two or more convolutional layers produces superior effects in object detection, deblurring and super-resolution, The author proposes a nested jump connection structure, which corresponds to the high-order residual learning of the conversion module in the defuzzification network.

Hongguang Zhang et al.~\cite{2019Deep} believe that multi-scale networks use a large number of training parameters to cause too long training time, and increasing the network depth of very low-resolution input does not seem to improve the deblurring performance, so they proposed a deep stacking Multi-patch network. The author proposes an end-to-end CNN hierarchical model similar to spatial pyramid matching. Compared with the scaling of the multi-scale model on the original size, the model directly cuts the original size picture into each small patch, and then the small In the next stage, the patches are spliced two by two into one large patch, which is carried out in stages. The model performs deblurring in fine to coarse meshes, thereby using multi-patch local to coarse operations. Each finer level functions as a residual, by contributing its residual image to a coarser level, allowing the network of each level to focus on a different blur scale.

Zamir SW et al.~\cite{2021Multi} proposed a multi-stage progressive image deblurring network.The author adds Channel Attention Block (CAB) to the feature extraction at each stage, which effectively extracts useful features. The channel attention module (CAM) has long shown an unparalleled role in various tasks of computer vision. For example, ~\cite{2020Learn}introduces channel attention to image classification tasks,~\cite{2020Weakly,2019A, 2019Incorporating}force the introduction of image segmentation tasks. Instead of simply cascading multiple stages, the author merges a supervisory attention module between every two stages. The supervisory attention module rescales the feature map of the previous stage, which can suppress the features with less information in the current stage and only allow useful features to propagate to the next stage. At the same time, the supervisory attention module provides a ground-truth supervisory signal, which is of great significance to the progressive image restoration at each stage. In addition, the author also proposed a cross-stage feature fusion mechanism, which makes the multi-scale features of the previous stage help enrich the features of the next stage, and also makes the network less susceptible to information loss. It is worth noting that the network architecture proposed in this article is not only effective in image de-modulation tasks, but also effective in image defogging, image de-raining, and image de-noising tasks.

\subsection{Generative adversarial deblurring network}
\label{Generative adversarial deblurring network}
In 2014, Ian Goodfellow~\cite{2014Generative} and his colleagues at the University of Montreal introduced Generative Adversarial Networks (GAN). This is a new method of learning the basic distribution of data, so that the generated artificial objects can achieve amazing similarity with real objects. The idea behind GAN is very intuitive: two networks of generator and discriminator play against each other. The goal of the generator is to generate an object (such as a photo of a person) and make it look the same as it really is. The goal of the discriminator is to find the difference between the generated result and the real image.Then, the method of adversarial learning is widely used in image tasks, such as the application of GAN network to segmentation tasks in ~\cite{yang2021task}.

\begin{figure}[htbp]
\centering
\includegraphics[height=10cm,width=10cm]{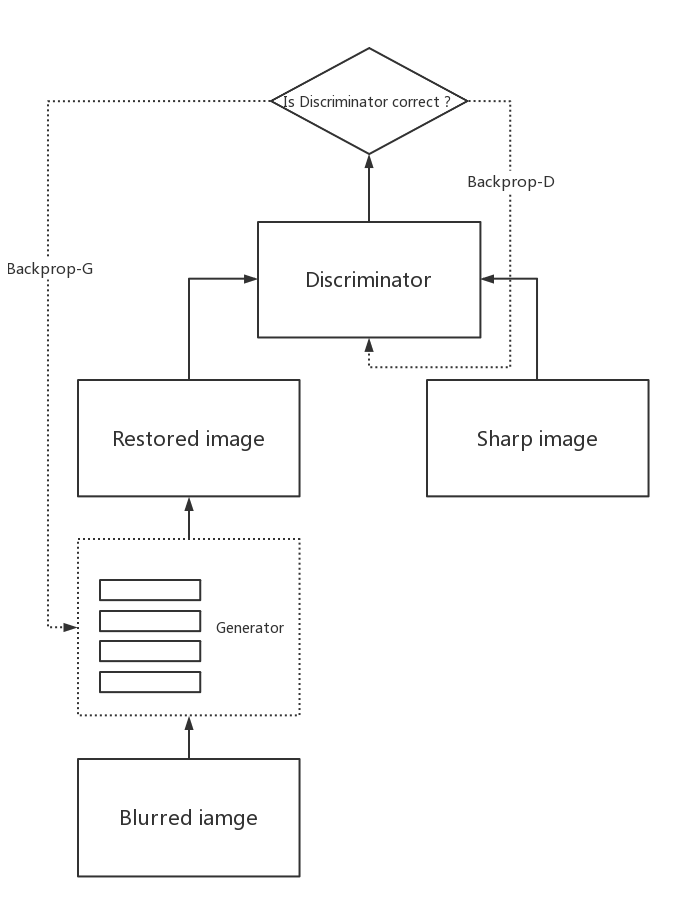}
\caption{GAN Model Framework Diagram}
\end{figure}

Orest Kupyn et al.~\cite{2018DeblurGAN} were inspired by the work on image super-resolution~\cite{2016Photo} and the image-to-image conversion in the generative adversarial network~\cite{2016Image}, and used image deblurring as a special case of this image-to-image conversion. Propose a de-fuzzy network that generates confrontation. The author constructed a generative adversarial network, trained a generator and a discriminator. In the training phase, the author not only trains the generator network but also introduces the discriminator network, and trains the two networks in a confrontational manner. The author hopes that through this alternate training method, the performance of the generator and the discriminator can be continuously improved, just like the fact that both sides can improve when competing with a good opponent. The author uses perceptual loss and confrontation loss as the total loss. The perceptual loss generally focuses on the recovery of general content and the confrontation loss focuses on the recovery of details. Compared with the L1 loss and L2 loss, the perceptual loss is a simpler L2 Loss is the loss of the feature map. In the test phase, a clear picture can be obtained by inputting the fuzzy picture into the trained generator network.

Orest Kupyn et al.~\cite{2019DeblurGAN} proposed a new framework of deblurgan-v2. Compared with~\cite{2018DeblurGAN}, its generator adopts a feature pyramid structure, and the discriminator uses a relativity discriminator. This feature pyramid model rebuilds a higher spatial resolution from the semantically rich layer through a top-down path. The horizontal connection between the compartments complements the bottom-up and top-down paths with high-resolution details and helps to locate objects. The author will take five final feature maps of different scales as output, and integrate all layers of information to improve the deblurring performance. Because the low-level feature semantic information is relatively small, but the target location is accurate; the high-level feature semantic information is richer, but the target location is relatively rough. In addition, although some algorithms use multi-scale feature fusion, they generally use the fused features to make predictions. The difference in FPN is that the predictions are performed independently at different feature layers, so the FPN structure undergoes different times of convolution and other processing. You can get different feature layer information.

\subsection{Unsupervised neural network}
\label{Unsupervised neural network}
Whether it is generating confrontational image deblurring networks or multi-scale image deblurring networks, there is a common problem-long training time and large amount of training data. Even if a powerful network model does not have sufficient training data, it cannot have a strong deblurring performance. As we all know, the traditional image deblurring algorithm can still achieve a good deblurring effect without a large amount of data support. How to bridge the gap between traditional algorithms and deep learning, build a bridge between traditional methods and deep learning methods, and build unsupervised neural networks has become a frontier issue. Dongwei Ren et al.~\cite{2020Neural} proposed the use of a deep and bright neural network, which can estimate the blur kernel and clear pictures under the condition of a known blur kernel size. Given a fuzzy picture and the size of the blur kernel, the author uses the known image size and the size of the blur kernel, and uses the two noises of the corresponding size as the input of the clear image generation sub-module and the blur kernel generation sub-module. By introducing DIP~\cite{2017Deep} and FCN respectively to capture the priors of clean images and blur kernels, the clear images and blur kernels can be recovered better. Use the obtained clear image and blur kernel and convolve the two to calculate the loss of the blurred image, and then update the parameters of the two sub-networks. After multiple iterations, a clear image generation sub-module and a blur kernel generation sub-module with excellent performance can be obtained. After a lot of experiments, the author found that the effect of joint optimization of the two sub-modules is significantly better than the effect of alternate optimization, and attributed it to the unconstrained blind deconvolution problem is a highly non-convex problem, so the alternate optimization method will make the optimization fall into Saddle point. This unsupervised method undoubtedly provides us with another brand new idea, an unsupervised idea, to solve the problem of image deblurring. Compared with the previous deep learning methods that require a large amount of data and a longer training time, it is undoubtedly a more advantageous party. However, this method must retrain the network for each single image and need a given blur kernel size, which appears to have certain limitations.

\tikzstyle{leaf}=[mybox,minimum height=1.2em,
fill=hidden-orange!50, text width=5em,  text=black,align=left,font=\footnotesize,
inner xsep=4pt,
inner ysep=1pt,
]

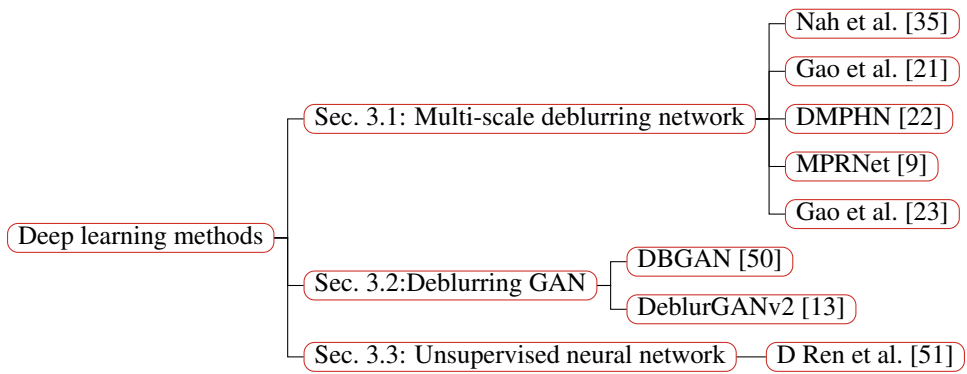
\begin{figure*}[thp]
 \centering
\begin{forest}
  forked edges,
  for tree={
  grow=east,
  reversed=true,  
  anchor=base west,
  parent anchor=east,
  child anchor=west,
  base=left,
  font=\normalsize,
  rectangle,
  draw=hiddendraw,
  rounded corners,
  align=left,
  minimum width=2.5em,
  inner xsep=4pt,
  inner ysep=0pt,
  },
    [Deep learning methods
        [Sec.~\ref{Multi-scale deblurring network}: Multi-scale deblurring network
        		[Nah et al.~\cite{2016Deep}
			]
			[Gao et al.~\cite{2018Scale}
			]
			[DMPHN~\cite{2019Deep}
			]
			[MPRNet~\cite{2021Multi}
			]
			[Gao et al.~\cite{2019Dynamic}
			]
        ]
        [Sec.~\ref{Generative adversarial deblurring network}:Deblurring GAN 
			[DBGAN~\cite{2018DeblurGAN}
			]
			[DeblurGANv2~\cite{2019DeblurGAN}
			]
        ]
        [Sec.~\ref{Unsupervised neural network}: Unsupervised neural network
			[D  Ren et al.~\cite{2020Neural}
			] 
        ]
    ]
\end{forest}
\caption{An image deblurring classification based on deep learning}
\label{deblurring_classification}
\end{figure*}

\section{Image deblurring dataset}
\label{Image deblurring dataset}
In this part, we will introduce some commonly used datasets in the field of image deblurring.

\subsection{GoPro dataset}
\label{GoPro dataset}
The GoPro dataset~\cite{2016Deep} is the most frequently used data set in deep learning methods. It includes 3214 pairs of clear pictures and fuzzy pictures, including 2103 pairs in the training set and 1111 pairs in the test set.

\subsection{RealBlur dataset}
\label{RealBlur dataset}
RealBlur dataset~\cite{2020Real} consists of two subsets that share the same image content, one of which is generated from the original image of the camera, and the other is generated from the JPEG image processed by the camera's ISP. Each subset provides 4,556 pairs of blurry and true sharp images of 232 pairs of low-light static scenes.

\subsection{HIDE dataset}
\label{HIDE dataset}
The HIDE dataset~\cite{2019Human} 8422 extensively annotates 65,784 human bounding boxes for clear pictures and blurred pictures. These images are carefully selected from 31 high fps videos, which not only include real outdoor scenes, but also different figures, poses and people appearing at different distances. The data set can also be divided into two categories, including long shots. (HIDEI) and ordinary pedestrians (close-up, HIDEII).

\subsection{Levin dataset}
\label{Levin dataset}
Levin dataset~\cite{2009Understanding} is the earliest fuzzy image dataset, which contains 32 grayscale pictures with 8 different blur kernels of 4 images and the corresponding clear pictures.

\tikzstyle{leaf}=[mybox,minimum height=1.2em,
fill=hidden-orange!50, text width=5em,  text=black,align=left,font=\footnotesize,
inner xsep=4pt,
inner ysep=1pt,
]

\begin{figure*}[thp]
 \centering
\begin{forest}
  forked edges,
  for tree={
  grow=east,
  reversed=true,  
  anchor=base west,
  parent anchor=east,
  child anchor=west,
  base=left,
  font=\normalsize,
  rectangle,
  draw=hiddendraw,
  rounded corners,
  align=left,
  minimum width=2.5em,
  inner xsep=4pt,
  inner ysep=0pt,
  },
    [Sec.~\ref{Image deblurring dataset}: Image deblurring dataset
        [Grayscale image dataset
            [Levin dataset~\cite{2009Understanding}     
            ]
        ]
        [RGB image dataset
            [RealBlur dataset~\cite{2020Real}
            ]
            [GoPro dataset~\cite{2016Deep}
            ]
            [HIDE dataset~\cite{2019Human}
            ]
	   ]
    ]
\end{forest}
\caption{Classification of datasets}
\label{org_survey_paper}
\end{figure*}
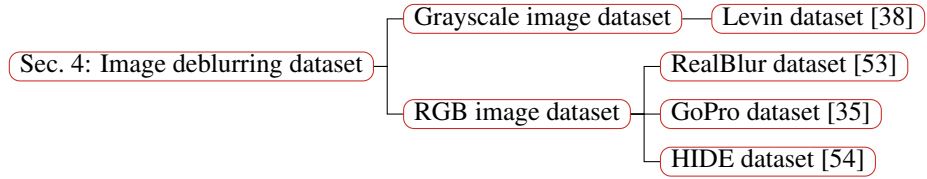

\section{Image quality evaluation}
\label{Image quality evaluation}
Image evaluation index has been widely used in other fields of image work.Image quality evaluation is one of the basic techniques of image processing~\cite{ma2016waterloo,wu2015blind,wu2018perceptually,wu2017blind,wu2020subjective,wu2014no}.Image quality evaluation is one of the basic techniques of image processing. It is an important indicator to reflect the quality of an image. It is mainly divided into subjective quality evaluation and objective quality evaluation. The subject of subjective quality evaluation is people, who score according to their viewing experience. This evaluation method is the most accurate and in line with human perception. The disadvantage is that the evaluation is troublesome and inefficient. Objective quality evaluation is based on image statistical information, so it has relatively many applications and can be divided into full reference, semi-reference and no reference.

\subsection{Subjective quality assessment}
\label{Subjective quality assessment}
The subjective quality scoring method is the most representative subjective evaluation method of image quality, which judges the image quality by normalizing the scores of the observers. The subjective quality scoring method can be divided into two types: absolute evaluation and relative evaluation.

The International Telecommunication Union (ITU) has proposed a variety of criteria for subjective evaluation methods, three of which are most commonly used, namely, the dual-stimulus damage grading method, the dual-stimulus continuous quality grading method, and the single-stimulus continuous quality grading method:Double stimulus damage grading method, given two sets of images of the original image (undistorted reference image) and the image to be tested (with certain distortion), compare the two sets of images to observe the damage of the image to be tested, according to the subjective quality of the image A 5-level scoring table to select the level of the image to be tested.The dual-stimulus continuous quality grading method is also given two sets of images, the difference is that the observer has no idea which is the reference image and which is the distorted image. The observer only needs to score the reference image and the test image separately according to the scoring table. Finally, the average value of the reference image and the image to be tested is calculated, and the difference between the two is calculated. The smaller the difference, the better the quality of the image to be tested.The single-stimulus continuous quality grading method is similar to the absolute subjective evaluation. The observer continuously scores the images to be tested according to the scoring table, and obtains the quality evaluation of the images to be tested according to the scoring and scoring time.

\subsection{Objective quality assessment}
\label{Objective quality assessment}
Due to the hot development of deep learning in the field of image deblurring, objective image quality evaluation based on full reference has become the mainstream. The objective evaluation method of image quality is to establish a mathematical model based on the subjective visual system of the human eye, and calculate the image quality through a specific formula. There are many traditional image objective quality evaluation indicators, among which MSE and PSNR are the most well-known.

\subsubsection{MSE}
\label{MSE}
Mean squared error is generally used to detect deviations between predicted and true values.
\begin{equation}\label{eq4}
MSE=\frac{1}{m}\sum_{1}^{m}\left ( y_{i}-\widehat{y_{i}} \right )^2
\end{equation}
where m represents the number of pixels, $y_{i}$ represents the real value of i pixel, and $\widehat{y_{i}}$ represents the predicted value of i pixel. The smaller the mean square error, the smaller the difference between the predicted value and the real value, and the more realistic the predicted image. However, in the image deblurring of today's deep methods, mse is often used as the loss during training, and PSNR and SSIM are more regarded as objective evaluation indicators for testing the deblurring effect.

\subsubsection{PSNR}
\label{PSNR}
Peak signal-to-noise ratio (PSNR) is an objective standard for evaluating images, and there are many application scenarios.
\begin{equation}\label{eq5}
PSNR=10 \log_{10}\left [ \frac{\left (2^{n}-1 \right )^{2}}{MSE}\right ]
\end{equation}
n is the number of bits per pixel, generally 8, that is, the number of pixel grayscales is 256. The unit of PSNR is dB, and the larger the value, the smaller the distortion. The smaller the MSE value, the closer the two pictures are, the larger the PSNR value, and the smaller the distortion. PSNR is the most common and widely used image objective evaluation index, but it is based on the error between corresponding pixels, that is, based on error-sensitive image quality evaluation. Since the visual characteristics of the human eye are not considered (the human eye is more sensitive to contrast differences with lower spatial frequencies, and the human eye is more sensitive to luminance contrast differences than chromaticity, the human eye's perception of an area will be affected by the results. The influence of the surrounding area, etc.), so the evaluation results are often inconsistent with people's subjective feelings.

\subsubsection{SSIM}
\label{SSIM}
In order to overcome the difference between objective indicators and subjective feelings, et al. proposed a new image quality evaluation indicator, namely structural similarity (SSIM). The system divides the tasks of similarity measurement into three comparisons: brightness, contrast, and structure.
\begin{equation}\label{eq6}
l\left ( x,y\right )=\frac{2{\mu _{x}}\mu _{y}+c_{1}}{\mu_{x}^{2}+\mu_{y}^{2}+c_{1}}
\end{equation}
\begin{equation}\label{eq7}
c\left ( x,y\right )=\frac{2{\sigma  _{x}}\sigma_{y}+c_{2}}{\sigma _{x}^{2}+\sigma _{y}^{2}+c_{2}}
\end{equation}
\begin{equation}\label{eq8}
s\left ( x,y\right )=\frac{{\sigma  _{xy}}+c_{3}}{\sigma _{x}\sigma _{y}+c_{3}}
\end{equation}
Generally take $c_{3}=\frac{c_{2}}{2}$, $\mu _{x}$ is the mean of $x$, $\mu _{y}$ is the mean of $y$, $\sigma_{x}^{2}$ is the variance of $x$, $\sigma_{y}^{2}$  is the variance of $y$, $\sigma _{xy}$ is the covariance of $x$ and $y$, $c_{1}=\left ( k_{1}L\right )^{2}$ and $c_{2}=\left ( k_{2}L\right )^{2}$ are two constants, $L$ is the range of pixel values, usually $k_{1}=0.01$, $k_{2}=0.03$ is the default value.So,
\begin{equation}\label{eq9}
SSIM\left ( x,y\right )=\left [ l\left ( x,y\right )^{\alpha } \cdot c\left ( x,y\right )^{\beta} \cdot s\left ( x,y\right )^{\gamma}\right ]
\end{equation}
Usually take $\alpha$,$\beta$,$\gamma$ as 1.The higher the SSIM value, the higher the similarity between the two images.

\section{Conclusion}
\label{Conclusion}
This article reviews the work of image deblurring. The traditional image deblurring method and the depth-represented image deblurring method are comprehensively summarized respectively. In the traditional image deblurring method, the method of maximum a posteriori and the variational Bayesian method are specifically introduced and analyzed; in the image deblurring method of depth representation, the multi-scale deblurring network and generation Confronting de-fuzzing networks and unsupervised neural networks. Finally, several widely used image deblurring datasets are also proposed. In general, this review provides considerable guidance for relevant researchers.

\bibliography{main1}
\bibliographystyle{IEEEtran}

\end{document}